\pdfoutput=1
\documentclass[sigconf]{acmart}

\usepackage{amsmath,amssymb,amsfonts}
\usepackage{algorithmic}
\usepackage{graphicx}
\usepackage{textcomp}
\usepackage{xcolor}
\usepackage{multirow}
\usepackage{dblfloatfix}
\usepackage{subcaption}
\usepackage{tabularx}

\acmConference[SEC '25]{The Tenth ACM/IEEE Symposium on Edge
Computing}{December 3--6, 2025}{Arlington, VA, USA}
\acmBooktitle{The Tenth ACM/IEEE Symposium on Edge Computing (SEC '25),
December 3--6, 2025, Arlington, VA, USA}
\acmDOI{10.1145/3769102.3770612}
\acmISBN{979-8-4007-2238-7/25/12}
\acmYear{2025}
\copyrightyear{2025}

\setcopyright{none}
\setcctype{by}

\begin{document}

\title{Elastoformer: Enabling Dynamic Adaptivity via Elastic Model Transformation}


\author{Sudaksh Kalra}
\affiliation{
  \institution{University of Amsterdam}
  \country{The Netherlands}
}
\email{s.kalra@uva.nl}

\author{Dolly Sapra}
\affiliation{
  \institution{University of Amsterdam}
  \country{The Netherlands}
}
\email{d.sapra@uva.nl}

\renewcommand{\shortauthors}{Kalra et al.}

\begin{abstract}
EdgeAI systems are increasingly employing computer vision applications to enable intelligent, on-device decision-making in real-time.  However, these deployments face highly dynamic operational conditions, with fluctuating constraints on latency, power availability, and memory resources. Deep Neural Networks (DNN), which follow fixed computational execution flows, lack the flexibility to adapt to such variability, resulting in inefficient and suboptimal performance in edge scenarios. This underscores the need for architectures that are not only efficient but also dynamically scalable at runtime. In this paper, we propose \textit{Elastoformer}: A framework that transforms conventional neural networks (NN) into \textit{Elastic} NN capable of real-time elastic inference. Unlike the conventional bag-of-models approach, which requires maintaining multiple independent models for different operating conditions, Elastoformer offers a single, modular solution that dynamically switches between multiple modes of operation at runtime, adapting efficiently to the changing computational budgets of edge devices without the overhead of managing separate models. Experiments reveal that our framework achieves up to 85\% reduction in computation FLOPs, 50\% reduction in latency and 76\% reduction in memory overhead, while showcasing the architecture agnostic nature of the framework across both Vision Transformers and CNNs. Our code is available at \url{https://github.com/sudaksh14/Elastoformer}.

\end{abstract}

\maketitle

\section{Introduction}
\label{intro}
Edge devices are increasingly being deployed for computer vision applications across diverse domains, including industrial automation, healthcare, autonomous vehicles and smart cities~\cite{10730112, chen_deep_2019}. These devices enable low latency, real-time decision making  without relying on cloud connectivity. They also process user data locally, making them well-suited for applications with strict security and privacy requirements. EdgeAI systems, in this context, refer to artificial intelligence solutions that operate directly on such edge devices, enabling on-device inference or learning under constrained computational, memory, and energy resources. Despite multiple benefits, EdgeAI systems face significant challenges~\cite{han_dynamic_2021, wang_not_2021, wen_adaptivenet_2023, han_legodnn_2021} due to limited computational resources, energy constraints, and storage limitations. Moreover, the availability of these resources is often dynamic, influenced by competing workloads and changing runtime conditions. To operate reliably under such variability, EdgeAI models must be capable of adapting their computational demands to the current state of the system.

To address these limitations, modern EdgeAI systems must incorporate more efficient neural networks that require reduced computational and memory resources. Furthermore, these networks should be capable of dynamically adapting their computational demands in response to changes in system state. The majority of the computer vision networks are monolithic in design with a fixed computational workflow. Traditional developments for improving computer vision performance typically involve increasing model complexity by adding parameters and raising the number of Floating Point Operations (FLOPs)~\cite{simonyan_very_2015,he_deep_2015,dosovitskiy_image_2021}. 

\begin{figure}[htbp]
\centering
\includegraphics[width=\columnwidth]{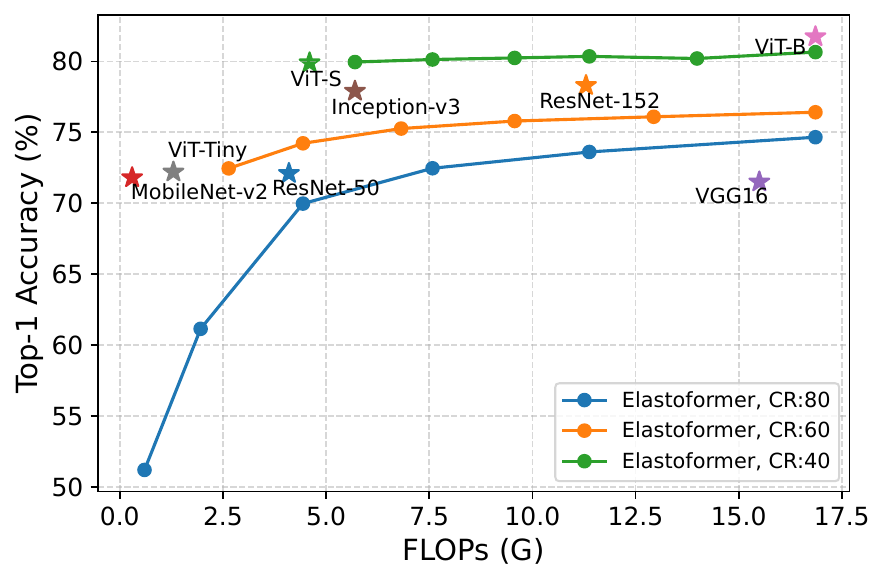}
\caption{Performance–efficiency Trade-off for Elastoformer compared to baseline vision models on ImageNet}
\label{imagenet_tradeoff}
\end{figure}

In this work, we propose {\em Elastoformer}, a dynamic neural network architecture that maintains competitive performance while adapting to continuously evolving system conditions of EdgeAI systems. Elastoformer provides a unified framework for transforming existing pre-trained models into elastic architectures capable of runtime adaptation. Fig.~\ref{imagenet_tradeoff} illustrates the trade-off between performance and resources (FLOPs) achieved by the proposed Elastoformer, in comparison with baseline vision architectures. While our method is motivated by the recent success of vision transformers~\cite{dosovitskiy_image_2021}, it is model-agnostic and applies equally to convolutional neural networks (CNNs), as also demonstrated in our experiments. 

At its core, Elastoformer incorporates a two step \textit{Compress} and \textit{Grow} framework to construct an elastic network which internally comprises of multiple \textit{Descendant Networks (DN)}. Each DN is capable of serving a different mode of operation for the EdgeAI system. This flexible design enables the instantiation of multiple runtime configurations from a single pre-trained model, supporting a wide range of compression levels to accommodate the dynamic and resource-constrained nature of edge environments.

Our work lies at the intersection of dynamic neural network and neural network compression. A wide range of techniques have been proposed to reduce the computational demands of neural networks, particularly the number of FLOPs, through various compression strategies. These include quantization~\cite{courbariaux2016binarizedneuralnetworkstraining, han_deep_2016} and pruning~\cite{han2015learningweightsconnectionsefficient,li_pruning_2017,molchanov_pruning_2017,fang_depgraph_2023}. Other  approaches include the design of compact architectures using either carefully crafted parameters~\cite{howard_mobilenets_2017} or automated neural architecture search (NAS) algorithms~\cite{elsken_neural_2019,sapra2020constrained}. While effective at reducing model size and inference cost, these approaches typically yield multiple independent neural networks or operating modes, each tailored to a fixed set of constraints. As a result, all variants must be stored simultaneously on the device, leading to significant memory overhead. Moreover, since there is no shared representation between modes, switching configurations requires paging entire model weight tensors in and out of memory at runtime~\cite{jiang_chameleon_2018,minakova_scenario_2022}. This lack of weight reuse not only increases memory consumption, particularly when the same layer structure appears across multiple modes, but also introduces additional latency during transitions, adversely affecting real-time performance.

This gap in existing approaches underscores the need for dynamic or elastic neural networks~\cite{han_dynamic_2021} that can adapt in real time to the changing compute, memory, and energy constraints inherent to EdgeAI systems. Prior works have explored input-adaptive models that reduce  computation based on the nature of incoming data~\cite{teerapittayanon_branchynet_2017, wang_not_2021, xu_lgvit_2023}. However, these approaches are generally tightly coupled to specific architectures and primarily exploit the nature of the inputs, offering limited benefits for broader system level variability in EdgeAI applications. 

Other efforts, particularly in the context of Vision Transformers (ViTs), employ specialized token merging or manipulation techniques to reduce inference costs~\cite{bolya_token_2023, rao_dynamicvit_2021, xu_evo-vit_2021}. While effective, these methods are static once trained, do not support runtime adaptation to varying resource budgets, and often require substantial retraining to recover baseline accuracy. Consequently, their applicability to dynamic and resource-constrained edge environments is limited.

We address said challenges by enabling elastic transformation of existing networks into multiple operational modes tailored to different compute and memory budgets. Elastoformer is specifically designed to support real-time adaptability with minimal memory and latency overhead, making it well-suited for edge computing scenarios where platform conditions may vary dynamically.

The key contributions of this work are as follows:
\begin{itemize}
    \item We introduce {\em Elastoformer:} an elastic vision transformer that dynamically adapts its size and computational complexity (FLOPs) based on real-time system constraints.
    \item We develop a general elastic transformation framework applicable to all neural architecture including CNNs, thus supporting broad deployment scenarios.
    \item Our approach minimizes memory overhead and transition latency by avoiding redundant storage of multiple independent models.
    \item We propose a novel weight-sharing growth mechanism that reuses parameters across Descendant Networks (DN), enabling efficient multi-mode operation with minimal retraining cost.
\end{itemize}

These contributions pave the way for more adaptable, resource-efficient neural architectures that align with the practical constraints and variability of real-world edge deployments.

\section{Background and Motivation}
The rapid proliferation of intelligent edge devices such as smartphones, IoT sensors, robots, autonomous drones and other embedded systems has catalyzed a significant shift in how AI is deployed. Instead of relying solely on centralized cloud infrastructure, there is a growing need to execute AI inference directly on edge devices, a paradigm referred to as Edge AI. This shift is driven by the demand for low-latency responses, reduced bandwidth consumption, enhanced data privacy and improved energy efficiency, all of which are critical in real-time and critical applications such as smart surveillance, autonomous navigation, industrial automation, and wearable health monitoring. We illustrate through a real world use case, how EdgeAI systems often operate in non-stationary environments with fluctuating resource availability and varying input complexity.

\subsection{Motivational Example}
Consider an autonomous drone equipped with a vision-based perception system for navigation, obstacle avoidance, and safe package delivery. Such a system typically integrates multiple vision tasks including image classification, object detection, and scene understanding, working in tandem with flight-critical subsystems to enable real-time situational awareness and safe decision-making.

Usually the runtime requirements of these systems vary considerably depending on the delivery scenario. Flying through dense urban areas imposes far higher performance demands than navigating open suburban or rural regions, and these demands can further fluctuate depending on temporal factors such as peak delivery hours versus off-peak periods or weekends. For instance, during urban peak hours, the drone must operate in highly dynamic and congested airspace, facing obstacles such as delivery vehicles, other drones, power lines, building edges, and temporary construction cranes. Environmental factors like strong winds, rain, fog, sun glare, or reflective surfaces, further degrade sensor reliability and challenge flight stability. These conditions demand high-accuracy, low-latency perception models capable of real-time obstacle detection, trajectory planning, and collision avoidance. Yet, onboard computational resources must still be shared with other critical subsystems, such as GPS navigation, communication, and payload management, which imposes tight constraints on the vision pipeline.

\begin{figure}[htbp]
\centering
\includegraphics[width=\columnwidth]{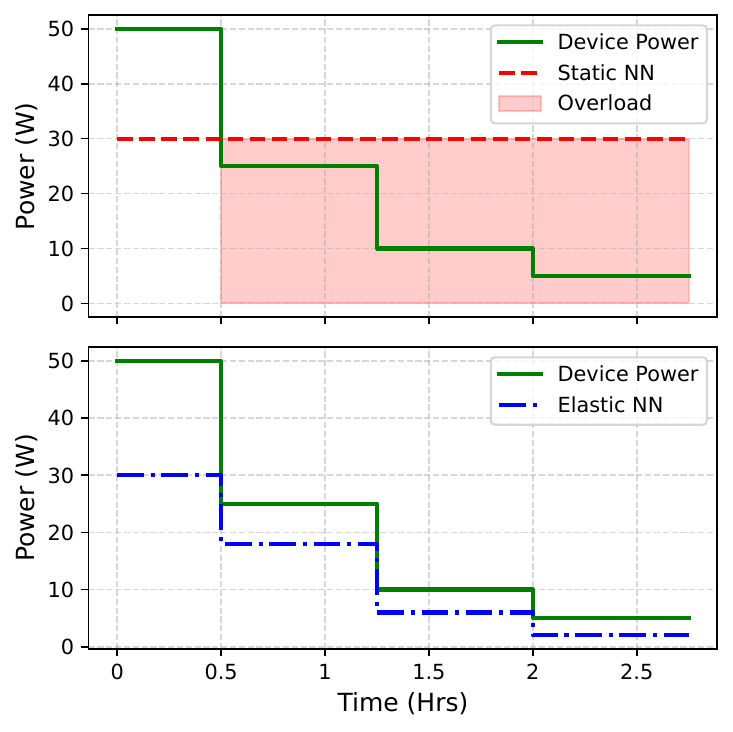}
\caption{Static (top) vs Elastic (bottom) NN under fluctuating device power budgets}
\label{motivation}
\end{figure}

To illustrate this challenge, Fig.~\ref{motivation} contrasts static and elastic neural networks under fluctuating device power budgets. In the static case (top), when available power drops (e.g., due to battery discharge), the DNN continues to demand fixed compute power, leading to increased latency or even system malfunction. In contrast, an elastic NN (bottom) scales its requirements in real time to remain within the available budget, enabling seamless perception-assisted flight without compromising safety.

In such dynamic environments, energy and memory resources for vision pipelines are inherently constrained by battery limitations, environmental conditions, temporal traffic fluctuations, and competing onboard processes. This variability underscores the need for computation-aware optimization strategies that adapt vision model execution to both the platform’s resource budget and the complexity of its operating context. Traditional neural networks with fixed architectures are optimized for average or worst-case scenarios, but this rigidity leads to inefficiencies: deploying a high-capacity urban-peak model in open rural regions wastes energy, while relying on a lightweight model during dense urban peak hours risks safety-critical failures.

This scenario is presented purely as a motivational example to illustrate variability in computational and energy requirements for drone deliveries and the potential benefits of adaptive neural networks. It does not represent the exact operating conditions of any specific commercial drone platform.

\subsection{Neural Network Elasticity}
To operate effectively under dynamic runtime constraints, modern AI systems, particularly those deployed at the edge, must go beyond static optimization and embrace neural network {\em Elasticity}. Elasticity in this context refers to the ability of a neural network to adapt its computational configuration, such as depth, width, or computational path, in response to runtime factors like latency, energy availability, memory constraints, or input complexity. This capability is especially critical in EdgeAI scenarios, where system conditions fluctuate frequently and over-provisioning for worst-case performance is often inefficient or unsustainable.
Elastic models are designed to offer a continuum of operational modes that can be selected dynamically at runtime, enabling the system to tailor its behavior to the current operating context. The overarching goal is to maintain a favorable trade-off between task performance and resource consumption, ensuring efficient and reliable inference even as execution conditions evolve.

A straightforward approach to achieve a flexible run-time behavior involves maintaining a set of independent pre-trained models, each optimized for a different point on the accuracy-efficiency spectrum. At runtime, the system switches between these models depending on the current resource availability or application demand~\cite{jiang_chameleon_2018,minakova_scenario_2022}. While conceptually simple, this bag-of-models approach incurs substantial memory overhead, which is especially problematic for edge devices with limited on-board storage. Moreover, runtime switching introduces non-trivial latency due to model loading, initialization, and parameter transfer overheads~\cite{fang_nestdnn_2018}.

In direct contrast, an elastic model does not rely on pre-loading multiple independent networks but rather encapsulates multiple execution paths or scalable configurations within a single shared architecture. This enables the system to make fast, low-overhead adjustments to inference cost while maintaining task accuracy under changing operating conditions. Elasticity can be manifested in various forms. A common strategy is to embed early-exit mechanisms~\cite{teerapittayanon_branchynet_2017,laskaridis_adaptive_2021} that allow inference to terminate at intermediate layers when sufficient confidence is reached. Another form involves scalable architectures that dynamically adjust width (e.g., number of active channels) or depth (e.g., number of layers processed), reducing the active portion of the network during execution~\cite{yu_slimmable_2018, yu_universally_2019, han_legodnn_2021, sun_steppingnet_2023}. More advanced designs integrate conditional computation, where only the most relevant parts of the model are activated per input, allowing for finer-grained control over compute and memory usage. While such approaches enable smoother adaptability without full model switching, many are based on custom-designed backbones or handcrafted layer structures, reducing compatibility with modern architectures, particularly those incorporating Multi-Head Attention (MHA) modules used in ViTs.

A more general solution involves using supernet-based frameworks, where a large over-parameterized network (a supernet) encodes a space of possible sub-networks. During deployment, the system dynamically selects sub-networks suited to the available resources~\cite{wen_adaptivenet_2023}. Despite their flexibility, supernet introduce significant challenges. The memory footprint is often $3\times$ larger than standard models, rendering them impractical for modern architectures like ViTs~\cite{dosovitskiy_image_2021}, which are already parameter-heavy.

The challenge in achieving effective elasticity lies in balancing flexibility with maintainability. Highly adaptable models must still ensure consistency in outputs, minimize switching overhead, and avoid significant accuracy degradation when operating in reduced modes. Furthermore, for modern architectures such as Vision Transformers, the design of elastic components must respect the structure of attention layers, which often resist naive pruning or modular scaling. Elasticity must be designed not as an afterthought but as an integral part of the model’s architecture, allowing it to degrade gracefully and recover efficiently based on runtime feedback.

In this context, {\em Elastoformer} is proposed as an elastic neural architecture specifically tailored for deployment on edge devices. Unlike approaches that rely on multiple model variants or heavyweight supernet, Elastoformer integrates elasticity directly into the backbone of a single model. These modes are not separate networks but part of a unified model structure, allowing Elastoformer to expand or shrink its computational graph with minimal switching overhead. This design enables the system to adapt seamlessly to fluctuating latency and energy constraints while preserving competitive accuracy.

\section{Related Works}
\subsection{Neural Network Compression}
Neural Networks have grown rapidly in terms of their computation and energy requirements in recent years. In practice, a larger network is able to generalize better because of more parameters and hence all research was focused on creating deeper and bigger architectures for increased performance~\cite{dosovitskiy_image_2021, he_deep_2015}. As more efficient deep learning applications are necessary for EdgeAI systems, the need for reducing computations of these networks also arise. Research have focused on methods to produce efficient networks by reducing the number of parameters, which in turn leads to reduced FLOPs. In this regard many research propose an optimized hand-crafted architectures~\cite{howard_mobilenets_2017,zhang_shufflenet_2017, touvron_training_2021} with reduced parameters and provide similar or improved performance than the existing state of the art architectures. However these approaches involve a custom architectures which are not general to adaptation to other networks and needs a lot of trial and error to estimate the best architecture combination. Quantization is another method which reduces the precision (e.g INT4 or INT8) of model parameters~\cite{courbariaux2016binarizedneuralnetworkstraining,han_deep_2016} to reduce the network FLOPs. However quantized networks require specialized hardware for actual energy and computational savings and thus are rendered incompatible with most GPUs. 

Pruning~\cite{10.5555/2969830.2969903,298572,han2015learningweightsconnectionsefficient,han_deep_2016} is another popular approach wherein the redundant parameters in the original network are removed to achieve smaller and more efficient networks without much loss in the performance. NN pruning could be broadly categorized into two types: A) Unstructured and B) Structured Pruning. Firstly, the unstructured approach involves selecting fine-grained independent neurons anywhere in the network, making those parameter weights as zero, thus leading to a sparse networks which are inefficient on most GPUs. Secondly we have structured approach which completely removes entire units or blocks of the network which leads to compact and dense networks with lower parameter count. The structured approach~\cite{fang_depgraph_2023,zhu_vision_2021} can be implemented on pre-trained models and could generate lighter, faster and efficient networks through little re-training after the pruned connections are removed.

\subsection{Dynamic Neural Networks}
Several existing studies have proposed dynamic architectures that adapt to evolving runtime conditions in edge environments. Slimmable Neural Networks~\cite{yu_slimmable_2018, yu_universally_2019} introduce a simple approach for training a single neural network capable of operating at multiple width configurations, thereby enabling real-time and adaptive trade-offs between accuracy and efficiency. Rather than training separate models for each width setting, a shared network is trained with switchable batch normalization layers to support multiple configurations within a unified architecture. However, this method relies on a custom backbone architecture that depends heavily on batch normalization, which poses a limitation for many state of the art architectures such as Vision Transformers (ViTs) that do not utilize batch normalization layers. 

AdaptiveNet~\cite{wen_adaptivenet_2023} enables elastic transformation by initializing a large-scale supernet through Neural Architecture Search (NAS) algorithms. This supernet is deployed on the target edge device, where a profiling algorithm is then employed to extract a sub-network tailored to the device’s resource constraints. While this method provides runtime elasticity, the supernet is approximately $3\times$ more overparameterized than the original architecture. This substantial overhead poses a significant bottleneck, particularly for modern architectures such as ViTs, which are already computationally intensive and challenging to deploy efficiently in edge environments.

\begin{figure*}[t]
\centering
\includegraphics[width=\textwidth]{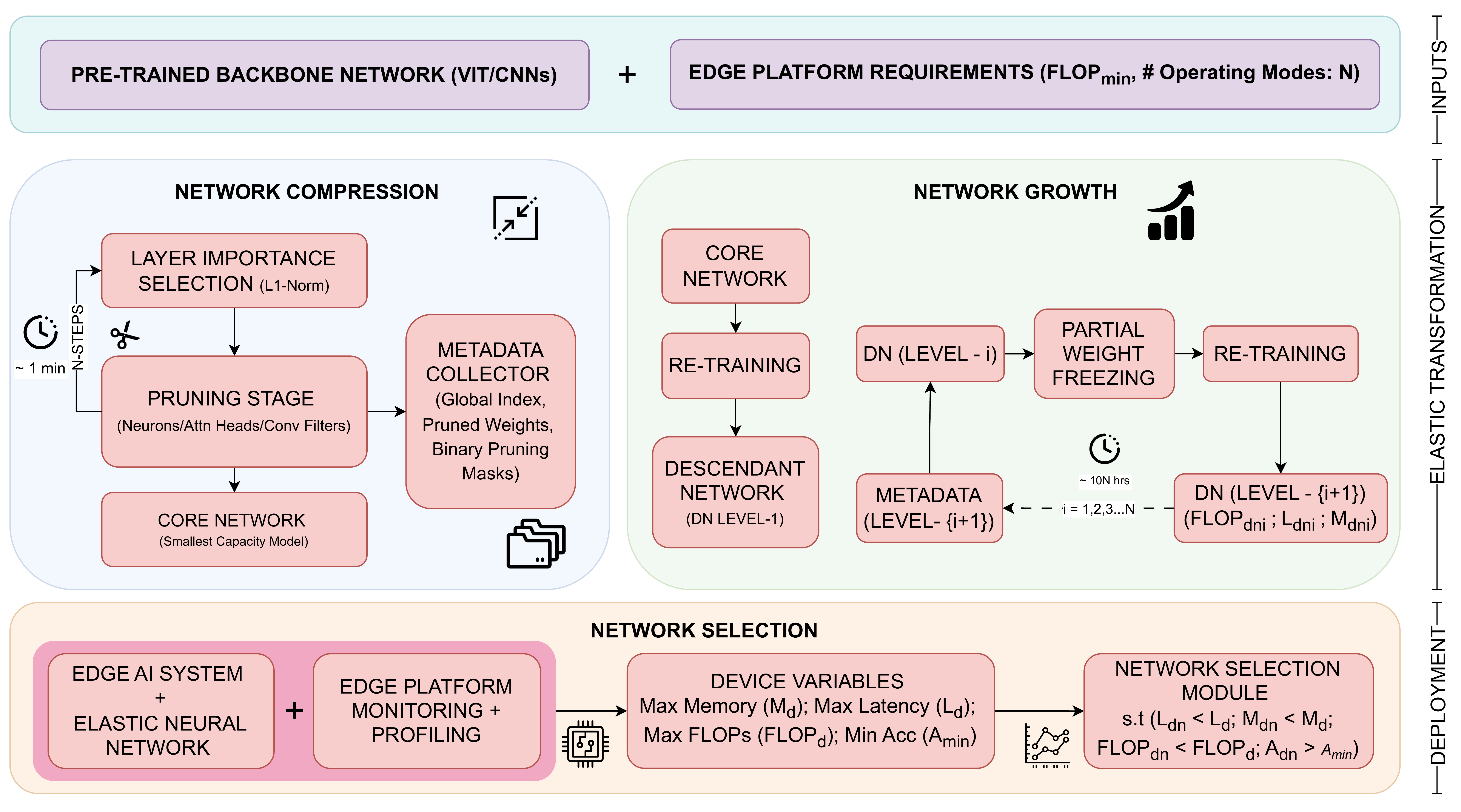}
\caption{Illustrative Schematic for Elastoformer Framework}
\label{scheme}
\end{figure*}

LegoDNN~\cite{han_legodnn_2021} introduces a block-grained scaling approach by training multiple interchangeable blocks for the same layers, enabling adaptation to dynamic edge environments. This method is designed to support modular architectures that can be reconfigured at runtime. However, the extent of compression achieved is limited, as the framework focuses solely on the intermediate layers that are not directly connected to the input or output, leaving the remainder of the network uncompressed. Furthermore, LegoDNN is constrained to convolutional neural network (CNN) backbones, as the block-grained design is incompatible with the multi-head attention (MHA) layers in transformer architectures. Consequently, this approach is not applicable to ViTs.

Early-exit networks represent another class of dynamic neural networks that incorporate multiple intermediate output layers throughout the backbone to enable early prediction and reduce overall computation~\cite{fang_flexdnn_2020, teerapittayanon_branchynet_2017, laskaridis_adaptive_2021}. These methods allow certain inputs to exit early based on confidence estimates, thereby offering dynamic inference paths. However, this introduces additional computational overhead due to the need for confidence evaluation at multiple points. Moreover, the decision to execute or skip modules is typically input-dependent, which limits the robustness of such approaches across the entire data distribution~\cite{han_dynamic_2021}. Certain methods, such as~\cite{fang_flexdnn_2020}, employ application-specific backbones tailored to particular downstream tasks e.g., continuous mobile vision but often fail to generalize across diverse architectures or datasets. Other approaches, such as~\cite{xu_lgvit_2023}, incorporate more complex exit branches using convolutional or attention mechanisms, further increasing the computational burden associated with dynamic inference and in turn the overall latency.

\section{Elastoformer Methodology}

We introduce {\em Elastoformer}, a framework that transforms pre-trained neural networks into elastic architectures capable of real-time adaptation to varying system conditions. {\em Elastoformer} enables the dynamic scaling of computation through a single unified model that supports multiple runtime configurations, each optimized for different resource constraints such as latency, memory, and power. 

As illustrated in Fig.~\ref{scheme}, the framework consists of three main stages: \textbf{(1) Input Specification}, where the pre-trained model and target edge constraints are defined; \textbf{(2) Elastic Transformation}, which constructs a family of sub-networks (termed \textit{Descendant Networks}) using a two-step \textit{Compress} and \textit{Grow} pipeline; and \textbf{(3) Deployment and Runtime Selection}, where the most suitable network is selected based on real-time system monitoring. The proposed design allows {\em Elastoformer} to operate under a continuum of performance-efficiency trade-offs while minimizing memory overhead and mode-switching latency.
The central component of {\em Elastoformer} is its two-stage transformation process, \textit{Compress} and \textit{Grow}, which converts a static pre-trained model into an elastic architecture capable of supporting multiple runtime configurations. In the \textit{Compress} stage, we progressively prune the model to generate smaller sub-networks that offer different trade-offs between computational cost and accuracy. In the subsequent \textit{Grow} stage, we incrementally restore pruned weights to produce a hierarchy of larger DN, each benefiting from shared parameters and warm-start initialization. These two stages work in tandem to produce an efficient and scalable set of models, all derived from a single original network. In this section, we describe each stage in detail.

\subsection{Neural Network Compression}
\label{compress}

We apply a structured pruning scheme to compress the original network into multiple progressively smaller variants as shown inside the blue block in Fig.~\ref{scheme}. Each pruning step removes a fraction of parameters from the attention and feed-forward layers, while also reducing hidden dimension sizes in the LayerNorm and Patch Embedding layers for greater compression.

Rather than relying on costly second-order gradient methods (e.g., Hessian-based saliency~\cite{10.5555/2969830.2969903, yang_global_2023}), we adopt a computationally efficient L1-Norm based saliency~\cite{li_pruning_2017,wang_why_2023}, which generalizes well across model types.

Let $N$ be the number of pruning rounds, $FLOP_B$ be the original FLOPs of the pre-trained network and $FLOP_{min}$ be the minimum available FLOPs for the edge system to run the smallest subnetwork (Core Network) under the specified latency constraints. We can estimate the the desired Compression Ratio, $CR$ and the pruning ratio per round, $p$ as described in~(\ref{eq:prune_ratio}).
\begin{equation}
\label{eq:prune_ratio}
\begin{aligned}
    CR = FLOP_{min} / FLOP_B \\
    p = 1 - (1 - CR)^{1/N}
\end{aligned}
\end{equation}

The encoder block of the ViT computes the Multi-Head Self-Attention (MHA) as described in~(\ref{eq:MHA}) following the original implementation~\cite{dosovitskiy_image_2021}, where an input image of size $(H \times W \times C)$ is first divided into $N = \tfrac{HW}{P^2}$ non-overlapping patches of size $(P \times P \times C)$, each linearly projected into embedding dimension $d$. Here, $Q, K, V \in \mathbb{R}^{n \times d}$ are the query, key, and value matrices derived from the patch embeddings, $W_i^Q, W_i^K, W_i^V \in \mathbb{R}^{d \times d_i}$ are the learned projections for the $i$-th head and $W^O \in \mathbb{R}^{hd_i \times d}$ is the output projection. Here, $n = N+1$ is the sequence length (including class token), $h$ is the number of attention heads and $d_i$ is the embedding dimension for head $h_i$, where $d_i = d/h$.

\begin{align}
\label{eq:MHA}
\text{MultiHead}(Q, K, V) &= \text{Concat}(\text{head}_1, \dots, \text{head}_h) W^O, \\
\text{head}_i &= \text{Attention}(Q W_i^Q, K W_i^K, V W_i^V), \nonumber \\
\label{eq:Attention}
\text{Attention}(Q, K, V) &= \text{softmax}\left(\frac{QK^\top}{\sqrt{d_i}}\right)V.
\end{align}

In each round, we prune the MHA projections as described in~(\ref{eq:MHA_pruning}) where \( M^\ast \in \{0,1\}^{d\times d} \) are binary pruning masks computed using an L1-norm-based saliency criterion and $\odot$ denotes element-wise multiplication. Specifically, for each weight matrix (e.g., \( W^Q \), \( W^K \), \( W^V \), or \( W^O \)), we compute the average absolute magnitude of each row. Rows with the lowest average L1-norms are assumed to have lower importance and are masked out (set to 0), while the remaining rows are retained (set to 1), enabling structured pruning of the attention projections.

\begin{equation}
\label{eq:MHA_pruning}
\begin{aligned}
    \tilde{W}^Q &= M^Q \odot W^Q, \quad
    \tilde{W}^K = M^K \odot W^K, \\
    \tilde{W}^V &= M^V \odot W^V, \quad
    \tilde{W}^O = M^O \odot W^O
\end{aligned}
\end{equation}

For the Feed-Forward (FF) layers as in~(\ref{eq:hidden}), we also prune the two-layer MLP as shown in~(\ref{eq:FF_pruning}):  

\begin{equation}
\label{eq:hidden}
\text{MLP}(x) = W_2 \cdot \phi(W_1 x + b_1) + b_2,
\end{equation}

where $x \in \mathbb{R}^{d}$ is the input, 
$W_1 \in \mathbb{R}^{d_{ff} \times d}$, 
$W_2 \in \mathbb{R}^{d \times d_{ff}}$, 
$b_1 \in \mathbb{R}^{d_{ff}}$, 
and $b_2 \in \mathbb{R}^{d}$.  
Here, $\phi(\cdot)$ denotes the non-linear activation (e.g., GELU). We keep $d_{ff} = 4\times d$ following~\cite{dosovitskiy_image_2021}.

\begin{equation}
\label{eq:FF_pruning}
\tilde{W}_1 = M_1 \odot W_1, 
\quad 
\tilde{W}_2 = M_2 \odot W_2,
\end{equation}

where $M_1 \in \{0,1\}^{d_{ff} \times d}$ and $M_2 \in \{0,1\}^{d \times d_{ff}}$ are binary masks, and $\odot$ denotes element-wise multiplication.

As mentioned earlier that our method is architecture agnostic and thus the pruning operation could be extended to CNNs as shown in~(\ref{eq:CNN_pruning}). For a convolutional layer with weight tensor $W_\text{conv} \in \mathbb{R}^{C_\text{out} \times C_\text{in} \times K \times K}$, where $C_\text{out}, C_\text{in}$ are the number output and input channels respectively and $K$ is the kernel size.

\begin{equation}
\label{eq:CNN_pruning}
\tilde{W}_\text{conv} = M_\text{conv} \odot W_\text{conv},
\end{equation}

Here $M_\text{conv} \in \{0,1\}^{C_\text{out} \times C_\text{in} \times K \times K}$ is a binary mask indicating active kernels. 

We store the pruned weights and corresponding masks after each round as pruning metadata. The smallest network obtained finally at the end of this stage is designated as the \textit{Core Network}.

\subsection{Neural Network Growth}
\label{growth}

To recover model capacity and construct higher-performing Descendant Networks (DNs), we incrementally reverse the pruning steps applied during compression as illustrated inside the green block in Fig.~\ref{scheme}. Starting from the \textit{Core Network}, the smallest and the most efficient model, we progressively build larger DNs by reintroducing a subset of the previously pruned weights. These weights are restored using the pruning metadata stored during the compression stage, which records their original positions and values.

To avoid redundancy and enable lightweight runtime switching between modes, we do not duplicate any weights. Instead, we employ a dedicated \textit{Weight Sharing} mechanism, as defined in~(\ref{eq:weight_sharing}). Specifically, $W_{L-1}$ denotes the fixed weights inherited from the previous DN, which are reused directly in the new network and remain frozen. In contrast, $W_l$ contains only the newly introduced parameters from the $L$-th growth iteration, which are updated during training. This selective update policy allows multiple DNs to coexist with minimal memory overhead while preserving consistency in the shared substructures. The composite weight matrix for the Level-$L$ DN, $W_L$, is given by~(\ref{eq:weight_sharing}).

\begin{equation}
\label{eq:weight_sharing}
\begin{aligned}
W_L &= W_l + W_{L-1}, \\
\text{where } 
W_l(i,j) &= 
\begin{cases}
w_{i,j}, & \text{if } (i,j) \in \text{Metadata}_{L} \\
0, & \text{otherwise}
\end{cases}
\end{aligned}
\end{equation}

\begin{figure}[htbp]
\centering
\includegraphics[width=\columnwidth]{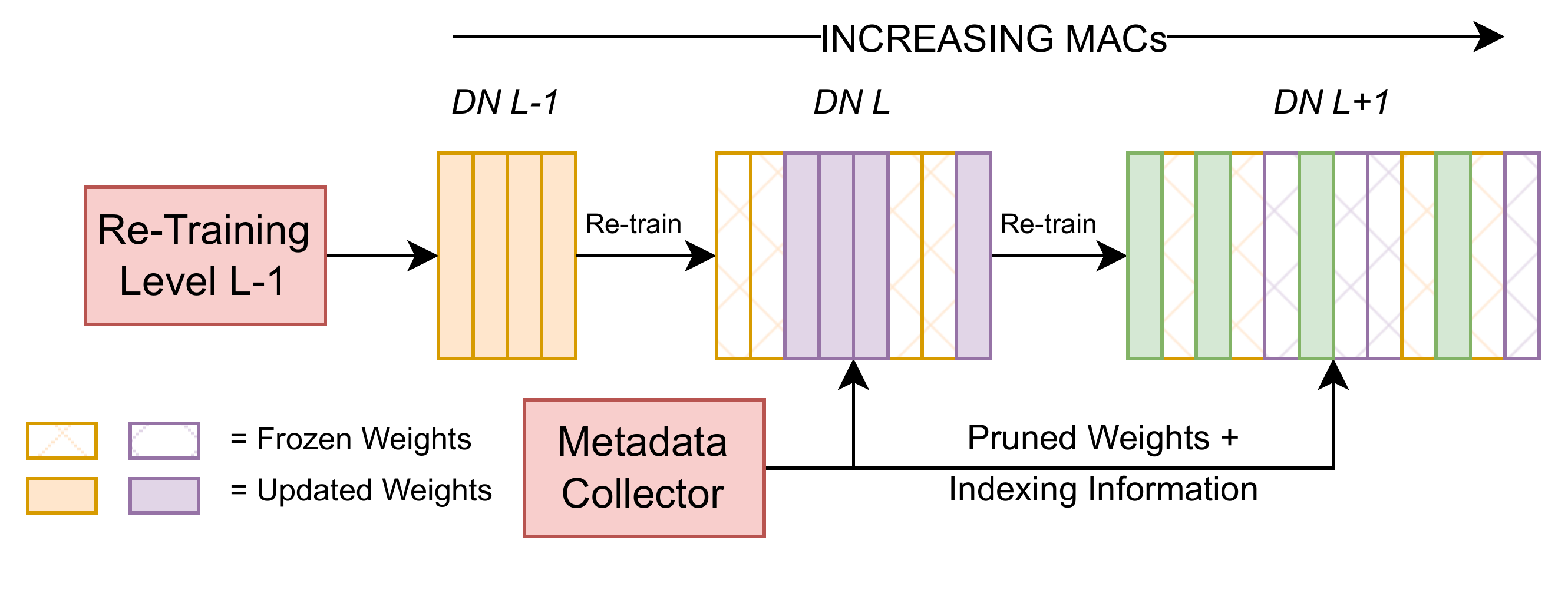}
\caption{Conceptual Illustration of Partial Weight Freezing for shared-weight elastic inference}
\label{freezing}
\end{figure}

As illustrated in Fig.~\ref{freezing}, each new DN at Level-$L$ is initialized by combining two components: (1) the preserved weights from the previous network that were not pruned, shown as white columns, and (2) the additional weights $W_l$ that were pruned in earlier iterations but are reactivated at iteration $L$, shown as dark shaded columns. This strategy leverages prior training while providing a warm start for the expanded network, thereby accelerating convergence and improving training efficiency.

To facilitate lightweight network growth without increasing memory overhead and to ensure stable retraining, we adopt two key strategies:

\subsubsection{Partial Weight Freezing}
To ensure the shared weights $W_{L-1}$ are not modified during training of the Level-$L$ DN, we apply partial weight freezing through gradient masking as in~(\ref{eq:weight_update}). Specifically, we use a binary mask $G$ to zero out the gradients of shared parameters during backpropagation:
\begin{equation}
W_L \leftarrow W_L - \eta \cdot \left(G \odot \nabla \mathcal{L}\right)
\label{eq:weight_update}
\end{equation}
where $G_{ij} = 0$ for all entries belonging to $W_{L-1}$ (shared weights), and $G_{ij} = 1$ otherwise. This guarantees that only the newly added weights are trainable, preserving compatibility with lower modes and avoiding unnecessary memory duplication.

\subsubsection{Selective Gradient Clipping}
\label{selective clipping}
To stabilize training, particularly for large datasets like Imagenet, gradient clipping is utilized to avoid exploding gradients~\cite{10.5555/3042817.3043083}. We apply a selective gradient clipping exclusively to the newly added weights to avoid the exploding gradient only for the parameters which are updated in training. Using the same binary mask $G$ as in~(\ref{eq:weight_update}), we restrict clipping to non-shared parameters. This prevents gradient explosion during re-training while maintaining the integrity of the frozen weights inherited from previous DNs.

\subsection{Elastic Neural Network}
\label{elastic_network}
The two-stage transformation process described above, \textit{Compress} and \textit{Grow}, collectively results in a modular, runtime-adaptive architecture that we term an \textit{Elastic Neural Network}. This architecture is not a single fixed model, but rather a collection of \textit{Descendant Networks (DNs)} derived from a common pre-trained backbone, each representing a distinct operating mode along a continuum of resource-accuracy trade-offs.

Each DN corresponds to a specific level of compression and is designed to meet varying deployment constraints. The smallest DN, the \textit{Core Network}, serves as the minimal resource configuration, while successive DNs incorporate more parameters and computational cost via progressive network growth. Importantly, all DNs share a unified parameter structure through the weight-sharing mechanism introduced during the growth stage. This ensures efficient memory utilization and enables seamless switching between modes with minimal overhead.
This elastic design allows the {\em Elastoformer} framework to dynamically adapt its capacity at runtime in response to the system's current state, offering real-time flexibility without the need to store or load multiple independent models. 

\subsection{System Monitoring and Network Selection}
\label{network_selection}

To enable runtime adaptivity, {\em Elastoformer} includes a lightweight system monitoring and network selection module deployed on the edge device as depicted inside the yellow block on the bottom of Fig.~\ref{scheme}. This module is responsible for selecting the most appropriate Descendant Network (DN) based on current resource availability and use case specific constraints.

During the design phase, each DN is profiled to obtain its corresponding accuracy (\%), latency (ms), memory footprint (in MB), and computational cost (in FLOPs). These values are stored as a look-up table on the device. At runtime, the edge system continuously monitors three critical parameters: available memory ($M_d$), allowable latency ($L_d$), minimum expected accuracy ($A_{min}$), and available power budget ($P_d$). The empirical relationship between power consumption and FLOPs is precomputed and stored as a lookup table to facilitate rapid decision-making. Given the current system state, our framework selects the DN with the lowest computational cost that satisfies all operational constraints as in~(\ref{eq:constraints}).

\begin{equation}
\begin{aligned}
\min_{\mathcal{DN}} \quad & \text{FLOPs}(\mathcal{DN}) \\
\text{subject to} \quad & \text{Latency}(\mathcal{DN}) \leq L_d \\
                        & \text{Memory}(\mathcal{DN}) \leq M_d \\
                        & \text{Accuracy}(\mathcal{DN}) \geq A_{min}
\end{aligned}
\label{eq:constraints}
\end{equation}

This formulation ensures that the selected network minimizes computation while adhering to the real-time performance and memory limits of the edge device. By combining metadata-aware scheduling with real-time profiling, the Elastoformer framework supports seamless and efficient transitions across elastic model configurations.

\section{Experimental Setup}
\subsection{Datasets}
The proposed method was evaluated on widely-used benchmark vision datasets: ImageNet~\cite{russakovsky_imagenet_2015}, CIFAR-10, and CIFAR-100~\cite{krizhevsky2009learning}. The CIFAR-10 dataset consists of 50,000 training images and 10,000 testing images, uniformly categorized into 10 classes, with all images having a resolution of $32\times32$. In contrast, the CIFAR-100 dataset contains the same number of training and testing samples as CIFAR-10, but is distributed across 100 fine-grained classes, making it a more challenging classification task. For both CIFAR datasets, standard data augmentation policies such as random cropping, horizontal flipping, and normalization were applied. The ImageNet-1K dataset, comprising 1,281,167 training images and 50,000 validation images spanning 1,000 object categories, was used for large-scale evaluation. In this case, extensive data augmentation strategies—including RandAugment and random erasing—were employed. Regularization techniques such as CutMix~\cite{yun_cutmix_2019} and Mixup~\cite{zhang_mixup_2018} were also utilized. Additionally, repeated augmentations were applied following the approach of~\cite{touvron_training_2021}.

\subsection{Backbone Networks and Baselines}
The Vision Transformer Base (ViT-B), comprising 12 layers, an embedding dimension of $d=768$, and a patch size of 16, as proposed in the original implementation~\cite{dosovitskiy_image_2021}, was selected as the pretrained backbone. Comparisons were conducted with several recent approaches for efficient vision transformers~\cite{bolya_token_2023, rao_dynamicvit_2021}. Additionally, comparisons were performed with Early-Exit ViT~\cite{xu_lgvit_2023}, which provides input-adaptive dynamic inference, and a recent approach~\cite{devvrit2024matformernestedtransformerelastic}, in which elasticity is introduced through slimmable feed-forward layers within the transformer blocks.

Further experiments were also conducted on CNNs (ResNet-50 and VGG-16), to evaluate the generalization of the approach beyond transformer-based architectures. The most recent works from each category, reporting results on the same backbone networks, were selected for benchmarking. In particular, the methods presented in~\cite{fang_flexdnn_2020} were chosen for early-exit networks. Approaches proposed in~\cite{wen_adaptivenet_2023, han_legodnn_2021, yu_slimmable_2018} were selected for their emphasis on adaptivity in dynamic neural networks.

\subsection{Evaluation Metrics}
The performance of the proposed approach was evaluated in terms of Top-1 accuracy with respect to changes in the total parameters of the descendant network and the variation in FLOPs. Latency of the DNs were also compared. In addition, total memory requirements for storing the networks were assessed, along with the average memory required for switching between different modes of operation. The total GPU hours consumed by the framework when applied to a completely new pre-trained network were also reported. The number of training hours was found to depend on the number of DNs required for a given use case, providing a representative estimate of the overall design time associated with the approach.

\subsection{Training Details}
Our framework was implemented on PyTorch~\cite{paszke2019pytorch} deep learning framework. All training procedures involved in the experiments were performed using eight NVIDIA GeForce RTX 3090 GPUs, each equipped with 24GB of memory. Each descendant network was trained for 50 epochs, with a linear warmup phase lasting 5 epochs. As previously mentioned, repeated augmentations were utilized, effectively increasing the training epochs by a factor of $3\times$. An initial learning rate of $3e^{-3}$ was used in conjunction with cosine annealing, unless specified otherwise. The AdamW optimizer was employed for all training processes, with a weight decay of 0.05 applied only to the core network. For all successive network re-training, weight decay was omitted. This was due to the incompatibility of weight decay regularization with the Partial Weight Freezing mechanism, as it also modified the gradients of shared parameters—an outcome that was undesirable for the proposed framework. The weight update involving weight decay is shown in~(\ref{eq:weight_decay}), where $w$ denotes the weight parameter, $\eta$ represents the learning rate, $\nabla_w \mathcal{L}$ indicates the loss gradient, and $\lambda$ denotes the weight decay rate. During training, no updates to the shared weights were desired and therefore, the weight decay parameter ($\lambda$) was set to zero.

\begin{equation}
    w \leftarrow w - \eta \cdot \nabla_w \mathcal{L} - \eta \cdot \lambda w
    \label{eq:weight_decay}
\end{equation}

A batch size of 128 was used for both training and evaluation. Stochastic depth~\cite{huang_deep_2016} was employed to regularize the training process, and selective gradient clipping was applied to prevent exploding gradients and promote faster convergence; however, this was done only on the parameters that were not shared with the previous DNs as discussed in Section~\ref{selective clipping}.

\section{Evaluation and Results}
We analyze our approach with focus towards performance and efficiency. We show the performance trade-off with reduced FLOPs, memory footprint and latency, resulting from our unique weight sharing mechanism.

\subsection{Performance vs Resource Trade-off}

\begin{figure}[b]
\centerline{\includegraphics[width=\columnwidth]{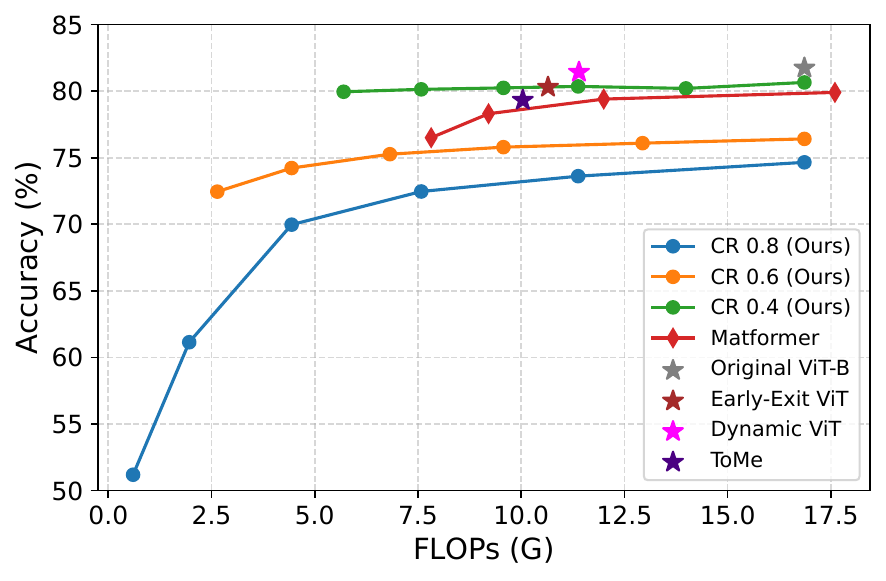}}
\caption{Performance vs. FLOPs trade-off under different compression ratios (CR $\in {0.4, 0.6, 0.8}$) on ImageNet}
\label{flops-vit}
\end{figure}

\begin{table*}[t]
\centering
\caption{Performance vs Resource Trade-off for ViT-B on ImageNet. Six Descendant Networks (DNs) are generated, where Level-1 is the smallest core network and Level-6 matches the original pre-trained network size. Base accuracy = 81.74\%, Base FLOPs = 16.86 G.}
\label{tab:performance}
\begin{tabular}{@{}c l c c c c c c@{}}
\toprule
\textbf{Compression Ratio} & \textbf{Metric} & \textbf{Level-1} & \textbf{Level-2} & \textbf{Level-3} & \textbf{Level-4} & \textbf{Level-5} & \textbf{Level-6} \\
\midrule
\multirow{2}{*}{0.40} & ACC \% (Top-1) & 79.95 & 80.13 & 80.24 & \textbf{80.35} & 80.20 & 80.13 \\
                       & FLOPs (G)      & 5.70  & 7.58  & 9.57  & 11.38          & 13.99 & 16.86 \\
\midrule
\multirow{2}{*}{0.60} & ACC \% (Top-1) & 72.45 & 74.22 & 75.26 & 75.79          & 76.09 & \textbf{76.41} \\
                       & FLOPs (G)      & 2.64  & 4.44  & 6.82  & 9.57           & 12.94 & 16.86 \\
\midrule
\multirow{2}{*}{0.80} & ACC \% (Top-1) & 51.19 & 64.14 & 69.97 & 72.46          & 73.61 & \textbf{74.65} \\
                       & FLOPs (G)      & 0.60  & 1.96  & 4.44  & 7.58           & 11.38 & 16.86 \\
\bottomrule
\end{tabular}
\end{table*}

We evaluate the multiple operational modes as shown in Table \ref{tab:performance} emerging from the elastic transformation of ViT on Imagenet. We vary the compression ratio (CR) across values in the set \{0.4, 0.6, 0.8\} to yield multiple performance baselines. We observe that our approach enables a flexible trade-off between computational cost (FLOPs) and classification accuracy (Top-1 \%). To highlight the generality and adaptability of our method, we select three representative compression ratios from this set and demonstrate how our approach supports a wide spectrum of accuracy-efficiency trade-offs as observed in Fig.~\ref{flops-vit}, thereby making it well-suited for deployment across devices with varying resource constraints. We observe that our approach performs at par with the original pre-trained network when utilizing the maximum Level-N descendant network, hence providing a fair trade-off. We observe FLOPs reductions by 85\% (CR = 0.6) while retaining more than 90\% of the peak original performance. With a lower overall compression (CR = 0.4), the smallest core network achieves up to 66\% reduction in FLOPs with an accuracy loss $\leq2\%$, further enhancing low-level (low FLOPs) DN efficiency.

We also compare our approach with an early-exit ViT~\cite{xu_lgvit_2023} as shown in Fig.~\ref{flops-vit}, here we need to note even if the performance is equivalent, early-exit networks offers a static memory footprint as the edge device needs to store the full architecture along with some additional exit-layer parameters which presents an additional overhead. We overcome this issue by providing actual memory saving as also discussed in Section \ref{memory overhead}.

Furthermore, we compared with other token merging approaches~\cite{rao_dynamicvit_2021, bolya_token_2023} which provide an efficient ViT with reduced or merged tokens. These methods provide a high performance with a reduced FLOPs requirement, however they fail to provide runtime adaptivity for fluctuating constraints which limits their application for dynamic edge deployments.

We also compare with another recent approach for elastic inference through nested subnetworks~\cite{devvrit2024matformernestedtransformerelastic}. We observe that our approach provides a better tradeoff than Matformer for the similar resource constraint.

We further extended our approach to convolutional neural networks (CNNs), including ResNet-50~\cite{he_deep_2015} and VGG-16~\cite{simonyan_very_2015}, as illustrated in Table~\ref{tab:cnn-perf}. We selected, CR = 0.5 for this experiment to generate six descendant networks (DNs), which offered a balanced trade-off between classification accuracy and computational complexity in terms of FLOPs. Our results reveal that the elastification process preserved the performance of the original pre-trained networks and, in the case of the CIFAR datasets, even led to improved accuracy. We observed a FLOPs reduction up to 75\%, representing a substantial efficiency gain that is particularly beneficial for EdgeAI deployment in resource-constrained environments.

\begin{table*}[t]
\centering
\caption{Performance vs Resource Trade-off for ResNet-50 and VGG-16 Descendant Networks (DNs) on benchmark vision datasets. Original network accuracy and FLOPs are indicated.}
\label{tab:cnn-perf}
\begin{tabular}{@{}l l l c c c c c c c@{}}
\toprule
\textbf{Architecure} & \textbf{Dataset} & \textbf{Metric} & \textbf{DN-1} & \textbf{DN-2} & \textbf{DN-3} & \textbf{DN-4} & \textbf{DN-5} & \textbf{DN-6} \\
\midrule
\multirow{6}{*}{\begin{tabular}[c]{@{}l@{}}ResNet-50 \\ (Orig FLOPs 4.12G)\end{tabular}}
& ImageNet (Orig Acc 80.35\%) & ACC (\%)  & 70.00 & 71.10 & 71.70 & 72.00 & 72.50 & \textbf{72.80} \\
&                              & FLOPs (G) & 1.07  & 1.51  & 2.03  & 2.65  & 3.33  & 4.12 \\
& CIFAR-10 (Orig Acc 94.37\%)  & ACC (\%)  & \textbf{95.08} & 94.99 & 94.91 & 94.89 & 94.92 & 94.85 \\
&                              & FLOPs (G) & 1.07  & 1.51  & 2.03  & 2.65  & 3.33  & 4.12 \\
& CIFAR-100 (Orig Acc 75\%)    & ACC (\%)  & 71.16 & 70.86 & 70.96 & 71.29 & \textbf{71.32} & 70.46 \\
&                              & FLOPs (G) & 1.07  & 1.51  & 2.03  & 2.65  & 3.33  & 4.12 \\
\midrule
\multirow{4}{*}{\begin{tabular}[c]{@{}l@{}}VGG-16 \\ (Orig FLOPs 15.53G)\end{tabular}}
& CIFAR-10 (Orig Acc 94.16\%)  & ACC (\%)  & \textbf{95.50} & 95.28 & 95.29 & 95.13 & 95.09 & 94.86 \\
&                              & FLOPs (G) & 3.96  & 5.60  & 7.59  & 9.91  & 12.51 & 15.51 \\
& CIFAR-100 (Orig Acc 74\%)    & ACC (\%)  & \textbf{77.78} & 76.58 & 77.14 & 76.62 & 76.77 & 76.88 \\
&                              & FLOPs (G) & 3.96  & 5.60  & 7.59  & 9.91  & 12.51 & 15.51 \\
\bottomrule
\end{tabular}
\end{table*}

\subsection{Performance vs Latency Trade-off}

\begin{figure}[htbp]
  \centering

  \begin{subcaptionbox}{Nvidia Jetson Orin\label{vit_latency_jetson_orin}}%
    {\includegraphics[width=0.48\textwidth]{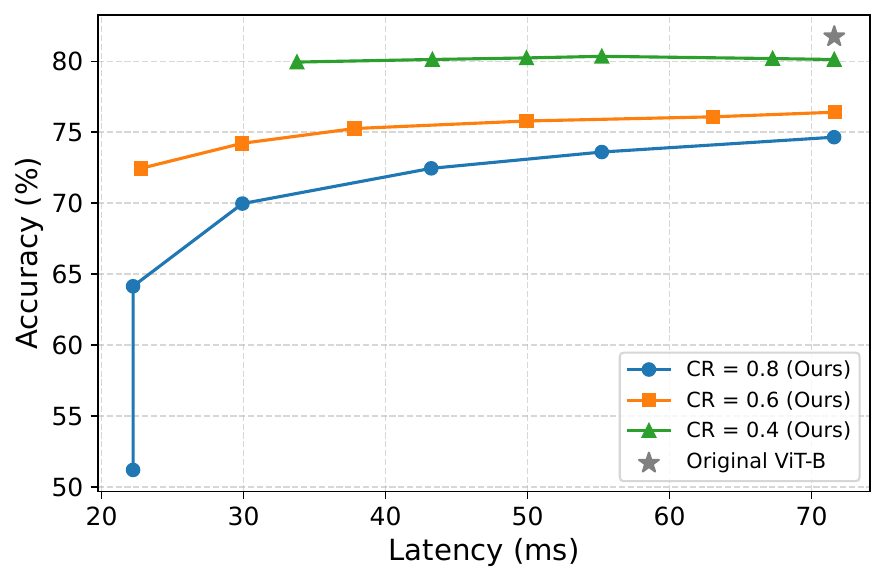}}
  \end{subcaptionbox}
  \hfill
  \begin{subcaptionbox}{Nvidia Jetson Nano\label{vit_latency_jetson_nano}}%
    {\includegraphics[width=0.48\textwidth]{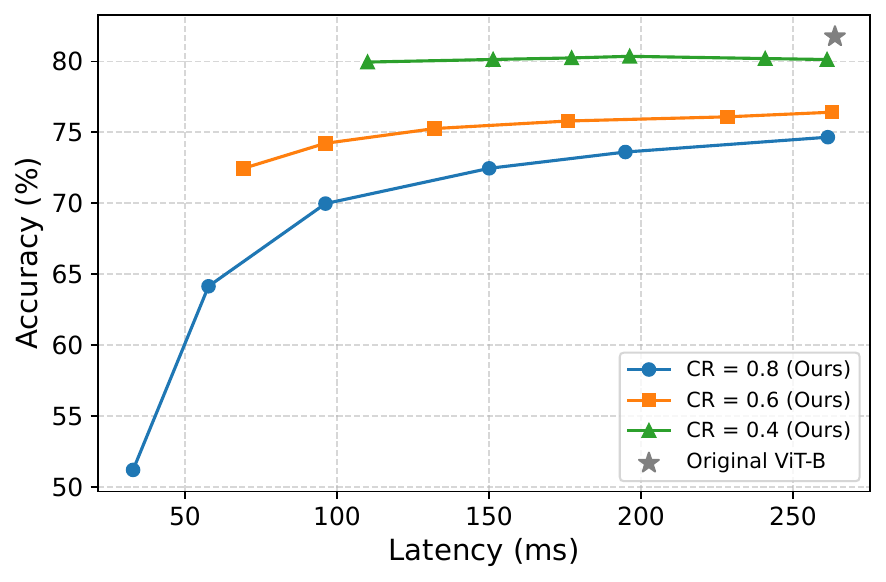}}
  \end{subcaptionbox}

  \caption{Performance vs. Latency trade-off under different compression ratios (CR $\in {0.4, 0.6, 0.8}$) on ImageNet}
  \label{latency-vit}
\end{figure}

In order to evaluate the effectiveness of the DNs achieved after the elastification, we evaluate the inference latency of the image classification on two platforms with different computation capabilities. We evaluated the latency of generated DNs on Nvidia Jetson Orin which supports an on-device Nvidia Ampere GPU~\cite{nvidia_jetson_orin} with 8GB memory and on Nvidia Jetson Nano~\cite{nvidia_jetson_nano} which also comes with an on-device Maxwell GPU having 4GB memory. We estimated the latency for 50 trials with a warmup of 10 trials for the GPU using a Batch Size of 1. We show in Fig.~\ref{latency-vit} how our elastic transformation can provide a competitive performance under variable latency constraints at the edge server.  

\begin{figure}[htbp]
\centerline{\includegraphics[width=\columnwidth]{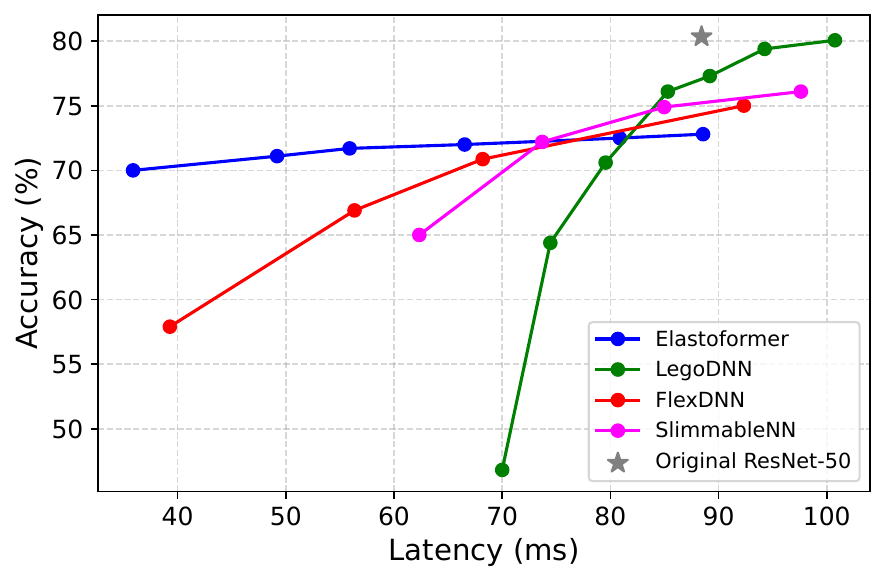}}
\caption{Performance vs. Latency trade-off of ResNet-50 on Jetson Nano, comparing Elastoformer against other adaptive neural network approaches}
\label{latency-resnet}
\end{figure}

We observe that the latency of the original pre-trained ViT-Base network is more than or equal to the biggest descendant network for all three compression ratio. As shown in Fig.~\ref{latency-vit}, the green curve corresponds to CR = 0.4. In this setting, Elastoformer achieves up to a 50\% reduction in latency across both edge platforms, while maintaining performance with less than a 2\% drop. We also evaluated the latency for Resnet-50 on Jetson Nano in Fig.~\ref{latency-resnet} to compare with some of the existing works. We observe in Fig.~\ref{latency-resnet} that our approach provides DNs which can work under much stringent latency requirements as compared to other approaches. Our DNs compromise on the accuracy when the FLOPs are on the higher side however other approaches need more FLOPs than the original network itself to maintain higher accuracy.

We also estimate the switching overhead latency when transitioning between different DNs on the Jetson Orin, observing a mean switching time of $\approx 50.4$ ms. This duration corresponds to approximately one inference cycle, indicating that the effective downtime is limited to a single cycle. As a result, our approach supports continuous inference with negligible impact on sustained throughput, even under frequent runtime adaptations. In contrast, conventional approaches that rely on reloading or retraining models typically incur substantially higher delays, making them impractical for real-time deployment on resource-constrained platforms.

\subsection{Memory Overhead Reduction}
\label{memory overhead}

We have showed in Table \ref{tab:memory}, the comparison of using our elastic transformation in comparison with AdaptiveNet~\cite{wen_adaptivenet_2023} and the bag of models approach. 

\begin{table*}[htbp]
\centering
\caption{Memory footprint comparison (including $\sim$1MB metadata per DN).}
\label{tab:memory}
\begin{tabularx}{\textwidth}{lXXX}
\toprule
Base Network (Orig. Size) & Elastoformer (Ours) & AdaptiveNet (SuperNet) & Bag of Models (Accum. DNs) \\
\midrule
ViT-B (330.3 MB)     & 336 MB   & —         & 843.17 MB \\
ResNet-50 (97.81 MB) & 103 MB   & 381.66 MB & 354.72 MB \\
\bottomrule
\end{tabularx}
\end{table*}

We observed the memory saving of 76\% in comparison with AdaptiveNet for ResNet-50 as per the results reported by them in their original work. In ViT, we notice a memory saving of 60\% in comparison to bag of models approach.

Our approach can lead to large reductions in memory as compared to existing approaches with a little overhead of $\approx1$MB per DN for storing the metadata over the original size of the network. This memory reduction is achieved because of the unique weight sharing mechanism between different DNs, which results in a single set of parameters for multiple modes of operation. This effect would be even more pronounced on larger backbone networks like Large Language Models.

\subsection{Elastic Transformation Design Time}
We show in Table~\ref{tab:gpu_hours} the total design time required by our approach for elastic transformation of the vision transformer for different number of modes of operation. For training the DNs we have used 8 Nvidia GeForce RTX 3090 GPU's each with a 24GB memory. We note the design time for multiple number of DNs and for two different datasets including Imagenet and CIFAR-100. 

\begin{table}[htbp]
\centering
\caption{Design time scaling for different Descendant Networks (DNs)}
\label{tab:gpu_hours}
\begin{tabular}{lcc}
\toprule
\textbf{\# DNs} & \multicolumn{2}{c}{\textbf{Total Design Time (Hrs)}} \\
\cmidrule(lr){2-3}
                 & \textbf{ImageNet} & \textbf{CIFAR-100} \\
\midrule
4  & 36  & 3.5 \\
6  & 61  & 5.4 \\
8  & 85  & 7.1 \\
11 & 120 & 8.4 \\
\bottomrule
\end{tabular}
\end{table}

We observe that the design time scales with both the target dataset and the number of DNs. Empirically, for ImageNet, the design time is $\approx10N$ hours, where $N$ is the number of DNs. For smaller datasets such as CIFAR-100, the design time is significantly lower, typically ($\leq 10\%$) of that for ImageNet, due to the reduced dataset size.

\subsection{Limitations and Future Work}
We observe that our core model capacity which is in practice defined through the minimum resources that the edge system could offer to run the vision application while meeting the latency requirement. We observe that the number of parameters in the core model mostly dictates the performance of the overall system including those of the successive descendant models. We note in Fig.~\ref{flops-vit} that with a higher compression ratio (CR = 0.8) the DNs are unable to reach a high performance while with a lower overall compression (CR = 0.4), the core model is able to regain back the oracle performance with only 33\% of the original FLOPs. We reason that the weight sharing mechanism is working for higher compression ratio which is lucrative for application in Large Language models which have high energy and computation demands and could prove to be a really efficient solution for those architectures. 

In future, we aim to develop an automated search algorithm that can determine the optimal number of descendant networks (DNs) and the corresponding compression ratios tailored to specific deployment requirements. Such an approach would enable efficient exploration of the accuracy-efficiency trade-off space with minimal human intervention. Additionally, incorporating a layer-wise distillation loss during the training of DNs could further enhance their performance by transferring knowledge more effectively from the original network to its compressed variants.

\section{Conclusion}
We presented {\em Elastoformer}, a novel elastification framework designed to transform static neural architectures, including Vision Transformers (ViTs) into elastic neural networks capable of adapting to dynamic resource constraints in real-time edge environments. The proposed framework generates multiple descendant networks (DNs) from a single pre-trained model, enabling a flexible trade-off between computational cost and predictive performance. Central to our approach is a novel weight sharing mechanism, which reduces memory overhead by up to 76\% compared to conventional methods, and lowers inference latency by up to 50\% relative to the original ViT. Our empirical evaluations demonstrate that Elastoformer achieves a competitive balance between efficiency and accuracy, achieving FLOPs reduction up to 85\% while retaining more than 90\% of the peak original performance. Furthermore, we show that our framework generalizes effectively to convolutional architectures such as ResNet-50 and VGG-16, underscoring its versatility. The Elastoformer pipeline offers an end-to-end transformation of pre-trained models into elastic networks with minimal retraining, making it a practical and scalable solution for deployment in resource-constrained edge AI systems.

\bibliographystyle{ACM-Reference-Format}
\bibliography{refs_acm}

\end{document}